\documentclass[11pt]{article} 

\newenvironment{klein}{\begin{footnotesize}}{\end{footnotesize}}

\usepackage{a4}
\usepackage{amssymb}
\usepackage{color}

\newif\ifpdf
\ifx\pdfoutput\undefined
\pdffalse 
\else
\pdfoutput=1 
\pdftrue
\fi

\ifpdf
  \usepackage[pdftex]{graphicx}
  \usepackage[pdftitle={XML framework for concept description and 
              knowledge representation},
              pdfauthor={Andreas de Vries},%
              pdfpagemode=FirstPage,%
              colorlinks=false,hyperindex=true,plainpages=false,%
  ]{hyperref}
  \DeclareGraphicsExtensions{.pdf,.jpg}
\else
  \usepackage{graphicx}
  \usepackage[pdftitle={XML framework for concept description and 
              knowledge representation},
              pdfauthor={Andreas de Vries},%
              pdfpagemode=FirstPage,%
              colorlinks=false,hyperindex=true,plainpages=false,%
              dvips=true,ps2pdf=true%
  ]{hyperref}
  \DeclareGraphicsExtensions{.eps}
\fi

\newcommand{\code} [1]{{\footnotesize\tt #1}}
\newcommand{\codlet} [1]{{\scriptsize\tt #1}}
\newcommand{\URL} [1]{\underline{\scriptsize\tt #1}}
\newcommand{\qq}[1]{\codlet{'\hspace*{-.7ex}'\hspace*{-.1ex}#1\hspace*{-.1ex}'\hspace*{-.7ex}'\hspace*{-.1ex}}}
\newenvironment{pgm} {\begin{tt}
                        \begin{footnotesize}
                          \begin{tabular}{l}
                          \\ }
                     {    \\
                        \end{tabular}
                        \end{footnotesize}
                       \end{tt} \\
                     }

\begin{document}
\title{XML framework for concept description
and knowledge representation}

\author{Andreas de Vries\thanks{	e-Mail:
	\href{mailto:de-vries@fh-swf.de}{de-vries@fh-swf.de}} \\
	{\footnotesize{\em 
	FH S\"udwestfalen University of Applied Sciences,
	Haldener Stra{\ss}e 182,
	D-58095 Hagen, Germany}}
}
\date{}
\maketitle

\begin{abstract}
An XML framework for concept description is given,
based upon the fact that the tree structure of XML implies the
logical structure of concepts as defined by attributional
calculus. Especially, the attribute-value representation
is implementable in the XML framework.
Since the attribute-value representation is an important
way to represent knowledge in AI, the framework offers
a further and simpler way than the powerful
RDF technology.
\end{abstract}

\paragraph{ACM:} 
I.7.2 XML, E.2 Object Representation, H.1.1 Information Theory, G.2.3


%

\section{Introduction}
Knowledge representation is
an important and wide area of Artificial Intelligence.
The simplest way to represent knowledge of
an object is the attribute-value representation. Here
an object is  characterized by its attributes, each
of which having a fixed range of value. A concept
then is a category of objects specified by a logical
combination of attribute values.

On the other hand
an XML document consists
of elements specified by attribute-value pairs
and nested in a hierarchical tree structure.
Hence the idea to apply XML to knowledge representation
via the attribute-value representation is straightforward.
The logical notions of rules and concepts are
rather naturally mapped into the XML framework, which
to demonstrate is the major intention of this paper.

The application of XML to concept description
has a lot of advantages.
Since XML is a universal
and web-based data format, it is appropriate for platform-independent
and world-wide use.
XML nowadays has become a widely accepted standard data interchange
technology
so that its general usability is guaranteed for a long time.

Another aspect is the power of XML to let check data 
consistency of instance documents against the corresponding
XML schema. This tears consistency checks apart from 
the applications processing the data.

The new XML framework supplements different
approaches towards
knowledge representation using RDF technology
to build a Semantic Web 
\cite{Berners-Lee-et-al-2001,Brickley-Guha-2000,%
Broekstra-et-al-2002,Broekstra-et-al-2001,%
FOL,Lassila-Swick-1999,McGuinness-et-al-2002,OKBC,OWL}.
But whereas RDF is the more powerful technology enabling
semantic links by an entity-relationship implementation, 
the XML framework is simpler and more appropriate
to a description of concepts based upon attribute-value representation
as well as to their storage and to data interchange.
The framework enables and simplifies 
worldwide data access
for subsequent applications, e.g.\ machine learning programs.

The present paper is organized as follows. In section
\ref{sec-xml} a sketchy overview of XML is given,
followed by a short introduction to knowledge representation 
and attributional calculus in section \ref{sec-attributes}.
The core of the paper is section \ref{XML-concept} modelling
concept description in XML, section \ref{emerald} provides 
Emerald's world as an example of this model. A short discussion
concludes the paper.

\section{XML}
\label{sec-xml}
Among other purposes,
XML has been constructed as a data interchange format.
By definition it is a textual markup language consisting
of \emph{elements} which are organized in a tree structure.
Syntactically, each element \textit{name} is 
opened by a start tag,
\verb?<?\textit{name}\verb?>?, and closed
by an end tag, 
{\footnotesize \verb?<?}\textit{/name}{\footnotesize \verb?>?}.
To reflect the tree structure, any element opened
after the start tag of a previous element must be closed
before the previous element is closed.

Any element can have $m$ children as well as $n$
attributes, $m,n\in\mathbb{N}_0$ (both $m$ and $n$ may
vanish). An attribute is written in the start tag of
the element with its value in quotation marks
({\footnotesize\verb?"?}), i.e.
\begin{center}
	{\footnotesize\verb?<?}%
	{\em name attribute%
	{\footnotesize\verb?="?}value}{\footnotesize\verb?">?} 
	\ \ldots \ \
	{\footnotesize\verb?</?}{\em name}%
	{\footnotesize\verb?>?}
\end{center}
If an element in an XML document has no child, it can be
written as 
{\footnotesize \verb?<?}\textit{name}{\footnotesize \verb?/>?},
i.e.
{\footnotesize\verb?<?}{\em name}%
{\footnotesize\verb?>?}
{\footnotesize\verb?</?}{\em name}%
{\footnotesize\verb?>?}
\ $=$ \
{\footnotesize\verb?<?}{\em name}%
{\footnotesize\verb?/>?}.
The possible elements of an XML document are declared
in its DTD \emph{(document type definition)} or, 
more generally, in its XML schema. An XML
schema by itself is an XML document with a predefined
element set. In particular, the root element of an XML
schema is {\footnotesize \verb?<xsd:schema>?} \ldots\
{\footnotesize \verb?</xsd:schema>?}.
Here \code{xsd} denotes the XML namespace (\code{xmlns})
of the 
W3C consortium,
\begin{center}
	{\footnotesize
	\verb?xmlns:xsd="http://www.w3.org/2001/XMLSchema"?}.
\end{center}
A typical XML schema looks like the following source code.

\begin{klein}
\begin{verbatim}
  <?xml version="1.0"?>
  <xsd:schema 
      xmlns:xsd="http://www.w3.org/2001/XMLSchema"
      targetNamespace="...">
    ...
  </xsd:schema>
\end{verbatim}
\end{klein}

\noindent
For details see \cite{Bray-et-al-2000,W3C-2001}.

\section{Knowledge representation and attributional calculus}
\label{sec-attributes}
Attributional calculus \cite{Michalski-2000} is a simple
description language whose representational power is between
propositional calculus and first order predicate logic.
It serves as a technique for knowledge representation.
In the sequel a set-theoretic version of 
attributional calculus
is presented, briefly compared to the standard VL1 notation.
 
Let $X$ be a given finite set of objects. An \emph{attribute}
then is a mapping $a$: $X$ $\to$ $W$, $x$ $\to$ $a(x)$, 
where $W$ is a given
finite set, the range of the attribute values of
the objects $x$ in $X$.
Therefore, $a(x)$ is the value of the attribute that 
object $x$ possesses.
Let the objects $x$ in $X$ be uniquely characterized 
by a finite set of attributes 
$A=\{a_1, \ldots, a_n\}$, where each attribute $a_i$ has a
fixed finite range $W_i=a_i(X)$, $i=1, \ldots,n$.
Then each attribute can be written as
\begin{equation}
	a_i: X \to W_i, 
	\quad
	x \mapsto a_i(x).
	\quad (i=1,\ldots,n)
\end{equation}
Thus by this choice of attributes 
an object $x$ is well distinguishable from
the others by all its attribute values $a_1(x)$, \ldots, $a_n(x)$.
In other words, each object $x$ corresponds uniquely to a vector of 
attribute values $w_1$, \ldots, $w_n$,
\begin{equation}
	x \cong (w_1, \ldots, w_n) 
	\in W_1 \times \ldots \times W_n.
\end{equation}
The attribute-value representation formalizes the information 
that we have about the
object set $X$ and its objects. Hence it is a precise
notion for (one kind of) knowledge representation.
 
The basic construct of attributional calculus is the
\emph{elementary selector}\index{elementary selector} 
$S(a,w)$ given for an attribute-value pair $(a,w)$ by
$S$: $\mathcal{D}$ $\to$ $2^{X}$, $(a,w)$ $\mapsto$ 
$S(a,w)$, with 
\begin{equation}
	S(a,w) = \{x \in X:\ a(x) = w \}.
	\label{def-selector}
\end{equation}
Here $2^X$ denotes the potential set of $X$, and
$\mathcal{D}$ is the attribute-value set,
$\mathcal{D}$ $=$ $\bigcup_{i=1}^n (\{a_i\} \times W_i)$.
In other words, $S(a,w)$ 
selects all objects whose attribute
$a$ has value $w$. Of course, this set might be empty.
An empty selector corresponds to a ``don't care.''
Often, a selector $S$ is written a bit sloppily as
\begin{equation}
	S \ = \ \{a = w\}.
	\label{a=w}
\end{equation}
This is justified by Boole's second law \cite[\S\,XII.1]{Birkhoff-1973}.
In VL1 notation \cite{Michalski-1973}, a general 
selector is written as
$[a \stackrel{r}{\to} w]$, where $\stackrel{r}{\to}$ is a relation satisfying
$\stackrel{r}{\to}$ $\in \{=$, $\ne$, $\geqq$, $>$, $\leqq$, $< \}.$
Since the range $W$ is finite, a general selector is
a disjunction of elementary selectors. For instance,
if $W=\{w_1, w_2, \ldots, w_m\}$, the selector 
$[a\ne w_1]$ is equivalent to 
$[a = w_2]$ $\vee$ $[a=w_3]$ $\vee$ \ldots $\vee$ $[a=w_m].$

A \emph{rule}\index{rule}
$R$ is an intersection of selectors $S_1$, \ldots, $S_m$,
\begin{equation}
	R = S_1 \cap \ldots \cap S_m,
	\label{rule}
\end{equation}
where selector $S_i$ corresponds to attribute $a_i$. Note that a
selector may be empty and can be omitted in this case.
A \emph{concept}\index{concept} $C$ is a union of
rules $R_1$, \ldots, $R_k$,
\begin{equation}
	C = R_1 \cup \ldots \cup R_k.
\end{equation}
In VL1 notation, an intersection corresponds to a conjunction,
and a union corresponds to a disjunction.
Analogously to the context of Boolean functions, 
we call this representation the
\emph{disjunctive normal form} \cite[\S\,III.5]{Birkhoff-1973}
of a concept.
In the next section it will be worked out that this
simple structure of concepts 
is naturally represented in XML.

\section{XML model of concepts}
\label{XML-concept}
The XML model of concepts relies on the fact that the
structure of XML implies the structure
of attributional calculus. Three implications are
immediately observed:
\begin{enumerate}
	\item a selector is representable by an attribute-value pair,
	cf.\ eq.\ (\ref{a=w});
	
	\item the intersection (conjunction) of selectors
	corresponds to a list of attribute-value pairs
	in a single element;
	
	\item the union (disjunction) of rules corresponds to 
	the creation of children to a parent element. 
\end{enumerate}
Thus the XML model of concepts is  
straightforwardly achieved: A concept can be considered
as an element which
has either no, one, or more rules as child elements; 
a selector is simply an attribute-value pair of a rule element.
For instance, a concept may be given as

\begin{pgm}
	\verb?<concept>?
	\\
	\phantom{+}
	 \verb?<rule? $a_i$\verb?="?$w_{ij}$\verb?"? 
	 $\ldots$ $a_k$\verb?="?$w_{kl}$%
	 \verb?"/>?
	\\
	\phantom{+}
	 \verb?<rule? $a_m$\verb?="?$w_{mn}$\verb?"? 
	 $\ldots$ $a_p$\verb?="?$w_{pq}$%
	 \verb?"/>?
	\\
	\verb?</concept>?
	\\
\end{pgm}
Here $a_i$ is attribute number $i$, $W_i$ is its range, 
and $w_{ij}\in W_i$ is one of its possible values.
Hence a concept can be graphically represented by the following
diagram.
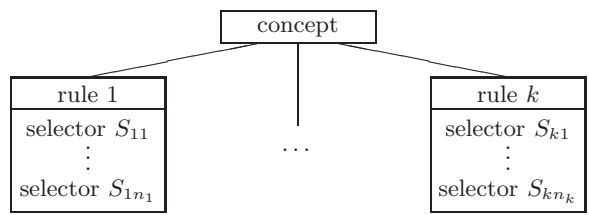
\begin{center}
\begin{klein}
\unitlength1ex
\begin{picture}(55,18)
	\put(28,17){\makebox(0,0){\framebox(15,3){concept}}}
	\put(7.5,12){\makebox(0,0)[t]{\framebox(15,2.95){rule 1}}}
	\put(7.5,8.97){\makebox(0,0)[t]{\framebox(15,10){}}}
	\put(7.5,7){\makebox(0,0){\makebox(15,10){selector $S_{11}$}}}
	\put(7.5,5){\makebox(0,0){\makebox(15,10){$\vdots$}}}
	\put(7.5,1){\makebox(0,0){\makebox(15,10)
	 {selector $S_{1n_1}$}}}
	\put(28, 5){\makebox(0,0){\makebox(15,3){\ldots}}}
	\put(48.5,12){\makebox(0,0)[t]{\framebox(15,2.95){rule $k$}}}
	\put(48.5, 8.97){\makebox(0,0)[t]{\framebox(15,10){}}}
	\put(48.5,7){\makebox(0,0){\makebox(15,10){selector $S_{k1}$}}}
	\put(48.5,5){\makebox(0,0){\makebox(15,10){$\vdots$}}}
	\put(48.5,1){\makebox(0,0){\makebox(15,10)
	 {selector $S_{kn_k}$}}}
	\put( 7.5,12.1){\line(5,1){16.5}}
	\put(28  , 7.4){\line(0,1){8}}
	\put(48.5,12.1){\line(-5,1){16.5}}
\end{picture}
\end{klein}
\end{center}
To enable this construct, the corresponding
XML schema for a concept has to be given 
as in the following source code.

\begin{klein}
\begin{verbatim}
  <?xml version="1.0"?>
  <xsd:schema
    xmlns:xsd="http://www.w3.org/2001/XMLSchema"
    targetNamespace=
      "http://www.math-it.org/xml/2002/concept.xsd"
    xmlns="http://www.math-it.org/xml/2002/concept.xsd"
    elementFormDefault="qualified"
  >
   <xsd:element name="concept">
    <xsd:complexType>
     <xsd:sequence>
      <xsd:element name="rule" minOccurs="0" 
                               maxOccurs="unbounded">
       <xsd:complexType>
        <xsd:attribute name="attribute_1" type="W_1"/>
        <xsd:attribute name="attribute_n" type="W_n"/>
       </xsd:complexType>
      </xsd:element>
     </xsd:sequence>
    </xsd:complexType>
   </xsd:element>
   <xsd:simpleType name="W_1">
    <xsd:restriction base="xsd:string">
     <xsd:enumeration value="w_11"/>
     <xsd:enumeration value="w_1m"/>
    </xsd:restriction>
   </xsd:simpleType>
   <xsd:simpleType name="W_n">
    <xsd:restriction base="xsd:string">
     <xsd:enumeration value="w_n1"/>
     <xsd:enumeration value="w_nk"/>
    </xsd:restriction>
   </xsd:simpleType>
  </xsd:schema>
\end{verbatim}
\end{klein}

\noindent
The names for the attributes, e.g.\ 
{\footnotesize\verb?attribute_1?}, 
as well as for the ranges (data types) of the values, e.g.\
{\footnotesize\verb?W_1?}, 
have to be adjusted appropriately.

This XML model can be easily extended to enable naming of
concepts or rules by an adding another
child to the \code{concept} element, or the
\code{rule} element, respectively.

\section{An example: Emerald's robots}
\label{emerald}
To illustrate the notion of the concept and its implementation
in XML, let us consider exemplarily
the world of Emerald's robots.\footnote{%
	\href{http://www.mli.gmu.edu/msoftware.html}
	{EMERALD} = Experimental Machine Example-based 
	Reasoning And Learning Disciple, see 	
	\href{http://www.mli.gmu.edu/msoftware.html}
	{\URL{www.mli.gmu.edu}};
	at this URL you also find Java-based animations 
	illustrating the system.%
}
It is a software system consisting of objects called
``robots.''
Each robot is described
by the values $w_1$, \ldots, $w_6$ of six attributes,
$w_i \in W_i$. 
The attributes with their ranges are listed in table
\ref{ml-tab-attributes}.
\begin{table}[ht] \centering
\begin{tabular}{l@{\ \ }l}
	\textbf{attributes} & \textbf{range of values} 
	\\
	\hline
	\codlet{headShape}
	  & $W_1$ $=$ \code{\{}\qq{round}, \qq{square}, 
				\qq{octagon}\code{\}} 
	\\
	\codlet{bodyShape}
	  & $W_2$ $=$  \code{\{}\qq{round}, \qq{square}, 
				\qq{octagon}\code{\}}
	\\
	\codlet{isSmiling}
	  & $W_3$ $=$ \code{\{}\qq{true}, \qq{false}\code{\}}
	\\
	\codlet{holding}
	  &  $W_4$ $=$ 
	    \code{\{}\qq{sword}, \qq{balloon}, \qq{flag}\code{\}} 
	\\ 
	\codlet{jacketColor}
	  & $W_5$ $=$ \code{\{}\qq{red}, \qq{yellow}, 
				\qq{green}, \qq{blue}\code{\}}
	\\
	\code{hasTie}
	  & $W_6$ $=$ \code{\{}\qq{yes}, \qq{no}\code{\}}
\end{tabular}
\caption{\label{ml-tab-attributes}%
	The robots' attributes and their ranges.%
}
\end{table}
A concept now is a specific description of a robot category.
For instance, 
\begin{center}
	$C = $ ``head is round and jacket is red, or head is square 
	and is holding a balloon''
\end{center}
is a concept. 
There are $3 \cdot 3 \cdot 2 \cdot 3 \cdot 4 \cdot 2 = 432$
different robot objects in this world, 84 of which belong to the
category $C$.
In our XML framework, the concept $C$ is implemented as

\begin{klein}
\begin{verbatim}
  <concept>
    <rule headShape="round" jacketColor="red"/>
    <rule headShape="square" holding="balloon"/>
  </concept>
\end{verbatim}
\end{klein}
The corresponding XML schema is given in the appendix;
it can also be found in the WWW at the URL

\href{http://www.math-it.org/xml/2002/emerald.xsd}
 {\URL{http://www.math-it.org/xml/2002/emerald.xsd}}

\noindent
It defines the concept structure as well as the
attribute ranges of the robots.

\section{Discussion}
In this paper an XML framework for concept description is
proposed. Concepts are expressed with the aid of
attributional calculus
as set-theoretic combinations of rules and selectors. 
An important observation is that
the structure of concepts
is implied by the structure of an XML document.
In particular, a selector is representable
by an attribute-value pair, a rule as
an intersection (conjunction) of selectors by
a list of attribute-value pairs in a
single element, and a union (disjunction) of rules
by a generation of children
of a parent element.
In this way, the XML framework for concept descriptions 
can be developed in a straightforward manner. 

As a 
consequence, the attribute-value representation in this
XML framework offers a route to represent knowledge,
distinct from similar but more powerful
approaches based, e.g., on the RDF technology 
\cite{Berners-Lee-et-al-2001,Brickley-Guha-2000,%
Broekstra-et-al-2002,Broekstra-et-al-2001,%
FOL,Lassila-Swick-1999,McGuinness-et-al-2002,OKBC,OWL}.
Since the framework refers to concepts 
in the sense of attributional calculus, 
it cannot represent \emph{all}
possible logical connectives. 
For instance, it does not provide
recursive structures 
(the mainstay of RDF where, e.g., the object of
an RDF property can itself have arbitrary properties)
or express inductive concepts which have only \emph{necessary}
conditions
(cf.\ the concept ``human'' and the classical
``featherless biped'' example due to Aristotle).
In addition, it has no quantification, negation, etc.,
although this could be implemented easily in a richer
schema for concept definitions, as is done in a wide range
of web-based knowledge representation languages such as 
OKBC \cite{OKBC}, DAML+OIL \cite{McGuinness-et-al-2002},
OWL \cite{OWL}, or full FOL \cite{FOL}.

Moreover, the attribute-value representation on which the
XML framework is based upon is only applicable to knowledge
systems consisting of solely \emph{finite} ranges of
attribute values.

However, the framework is still rich enough to
tackle problems of machine learning. 
Here the emphasis is laid
upon the representation of learning examples,
which are completely determined by their attributes as well as
an additional Boolean flag indicating whether they are
positive or negative. 
Thus the framework enables to store concepts in XML
documents and to let them be further processed by concept
learning algorithms. Reconciliation of a concept
document with its corresponding XML schema by 
standard XML parsers can be used as a data consistency
check, independently from subsequent processing programs.
The example of Emerald's robots indicates a perspective how this can
be done.

Since moreover
XML is a universal data interchange format, 
a web-based storage of concepts 
could serve as a germ of a standardized
world-wide knowledge database platform.

\section*{Appendix: XML schema of Emerald's world}
\begin{klein}\begin{verbatim}
 <?xml version="1.0"?>
 <xsd:schema
   xmlns:xsd="http://www.w3.org/2001/XMLSchema"
   targetNamespace=
         "http://www.math-it.org/xml/2002/emerald.xsd"
   xmlns="http://www.math-it.org/xml/2002/emerald.xsd"
   elementFormDefault="qualified"
 >
  <xsd:element name="emerald">
   <xsd:complexType>
    <xsd:sequence minOccurs="0" maxOccurs="unbounded">
     <xsd:element name="concept" minOccurs="0" 
                                 maxOccurs="unbounded">
      <xsd:complexType>
       <xsd:sequence>
        <xsd:element name="rule" minOccurs="0" 
                                 maxOccurs="unbounded">
         <xsd:complexType>
          <xsd:attribute name="headShape" type="HeadShape"/>
          <xsd:attribute name="bodyShape" type="BodyShape"/>
          <xsd:attribute name="isSmiling" type="IsSmiling"/>
          <xsd:attribute name="holding" type="Holding"/>
          <xsd:attribute name="jacketColor" type="Color"/>
          <xsd:attribute name="hasTie" type="HasTie"/>
         </xsd:complexType>
        </xsd:element>
       </xsd:sequence>
      </xsd:complexType>
     </xsd:element>
    </xsd:sequence>
   </xsd:complexType>
  </xsd:element>
  <xsd:simpleType name="HeadShape">
   <xsd:restriction base="xsd:string">
    <xsd:enumeration value="round"/>
    <xsd:enumeration value="square"/>
    <xsd:enumeration value="octagon"/>
   </xsd:restriction>
  </xsd:simpleType>
  <xsd:simpleType name="BodyShape">
   <xsd:restriction base="xsd:string">
    <xsd:enumeration value="round"/>
    <xsd:enumeration value="square"/>
    <xsd:enumeration value="octagon"/>
   </xsd:restriction>
  </xsd:simpleType>
  <xsd:simpleType name="IsSmiling">
   <xsd:restriction base="xsd:boolean">
    <xsd:pattern value="true"/>
    <xsd:pattern value="false"/>
   </xsd:restriction>
  </xsd:simpleType>
  <xsd:simpleType name="Holding">
   <xsd:restriction base="xsd:string">
    <xsd:enumeration value="sword"/>
    <xsd:enumeration value="balloon"/>
    <xsd:enumeration value="flag"/>
   </xsd:restriction>
  </xsd:simpleType>
  <xsd:simpleType name="Color">
   <xsd:restriction base="xsd:string">
    <xsd:enumeration value="red"/>
    <xsd:enumeration value="yellow"/>
    <xsd:enumeration value="green"/>
    <xsd:enumeration value="blue"/>
   </xsd:restriction>
  </xsd:simpleType>
  <xsd:simpleType name="HasTie">
   <xsd:restriction base="xsd:string">
    <xsd:enumeration value="yes"/>
    <xsd:enumeration value="no"/>
   </xsd:restriction>
  </xsd:simpleType>
 </xsd:schema>  
\end{verbatim}
\end{klein}

\begin{footnotesize}

\end{footnotesize}
\end{document}